\documentclass[a4paper,11pt]{article}

\usepackage{graphicx}
\usepackage[margin=1in]{geometry}
\usepackage{amsmath}
\usepackage{url}

\usepackage{multirow}
\usepackage{wrapfig}
\usepackage{svg}
\usepackage{caption}
\usepackage{subcaption}
\usepackage{cancel}

\begin{document}

\title{Path Space Partitioning and Guided Image Sampling for MCMC}

\author{
    Thomas Bashford-Rogers\\
    University of Warwick \\
    \texttt{thomas.bashford-rogers@warwick.ac.uk}
    \and
    Luis Paulo Santos\\
    Universidade do Minho \\
    \texttt{psantos@di.uminho.pt}
}

\maketitle

\begin{abstract}
  Rendering algorithms typically integrate light paths over path space. However, integrating over this one unified space is not necessarily the most efficient approach, and we show that partitioning path space and integrating each of these partitioned spaces with a separate estimator can have advantages. We propose an approach for partitioning path space based on analyzing paths from a standard Monte Carlo estimator and integrating these partitioned path spaces using a Markov Chain Monte Carlo (MCMC) estimator. This also means that integration happens within a sparser subset of path space, so we propose the use of guided proposal distributions in image space to improve efficiency. We show that our method improves image quality over other MCMC integration approaches at the same number of samples.
\end{abstract}

\begin{figure}[ht]
\setlength\tabcolsep{0pt}
\renewcommand{\arraystretch}{0}
\begin{center}
\begin{tabular}{ccc}
\includegraphics[width=0.3\textwidth]{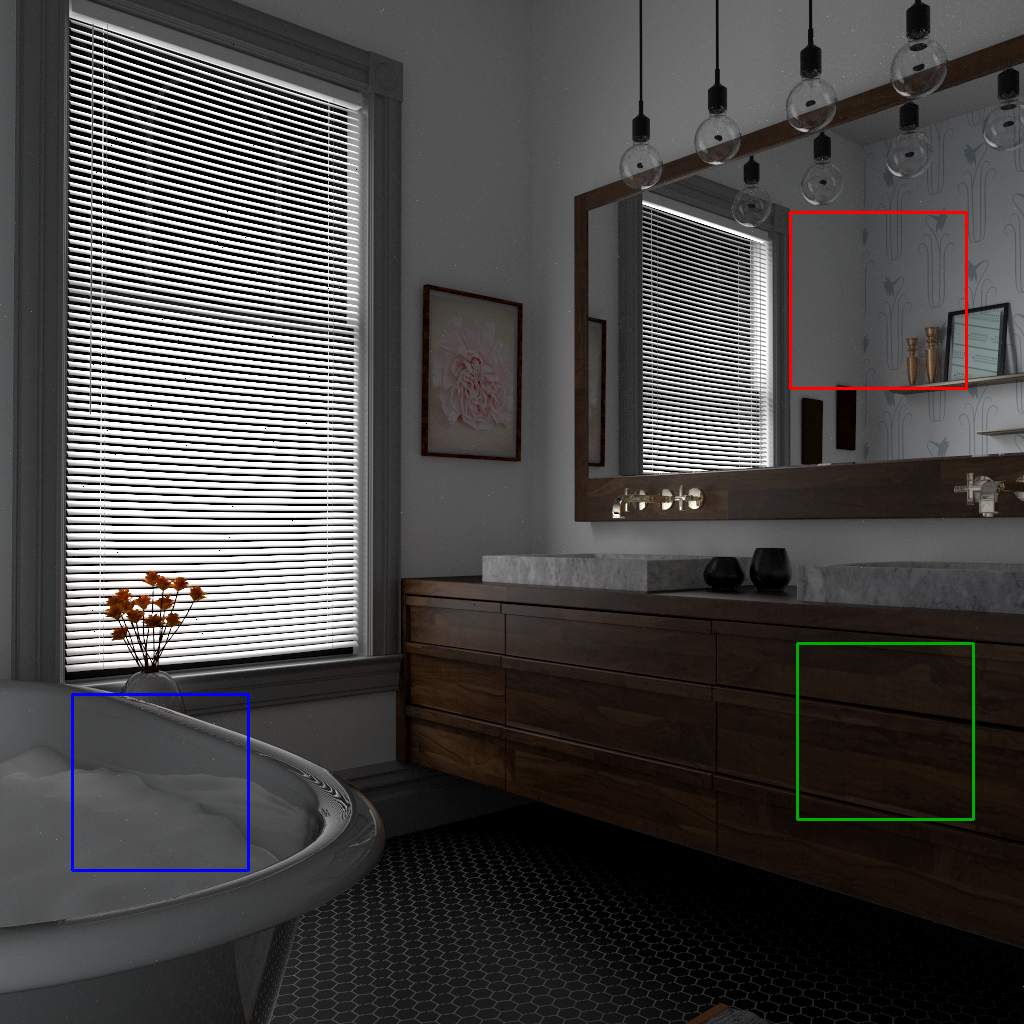} & \includegraphics[width=0.3\textwidth]{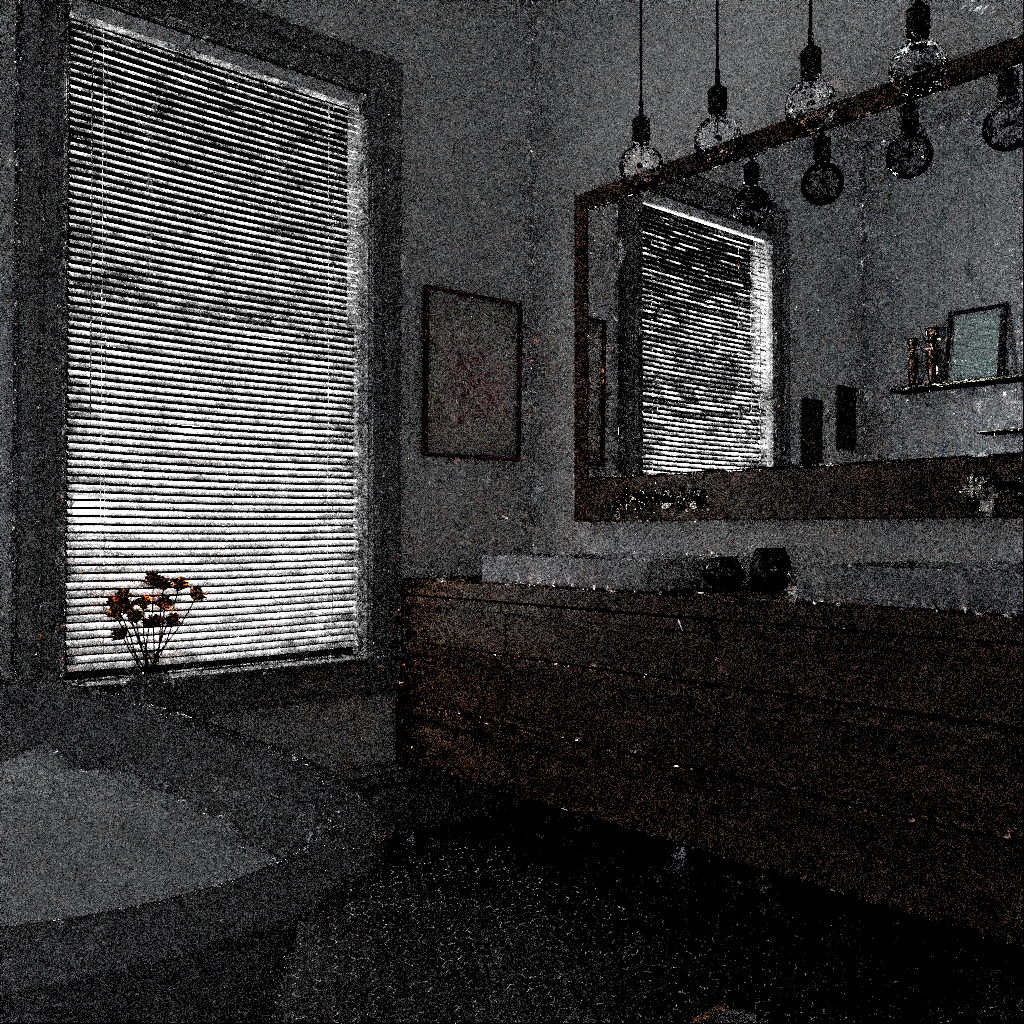} & \includegraphics[width=0.3\textwidth]{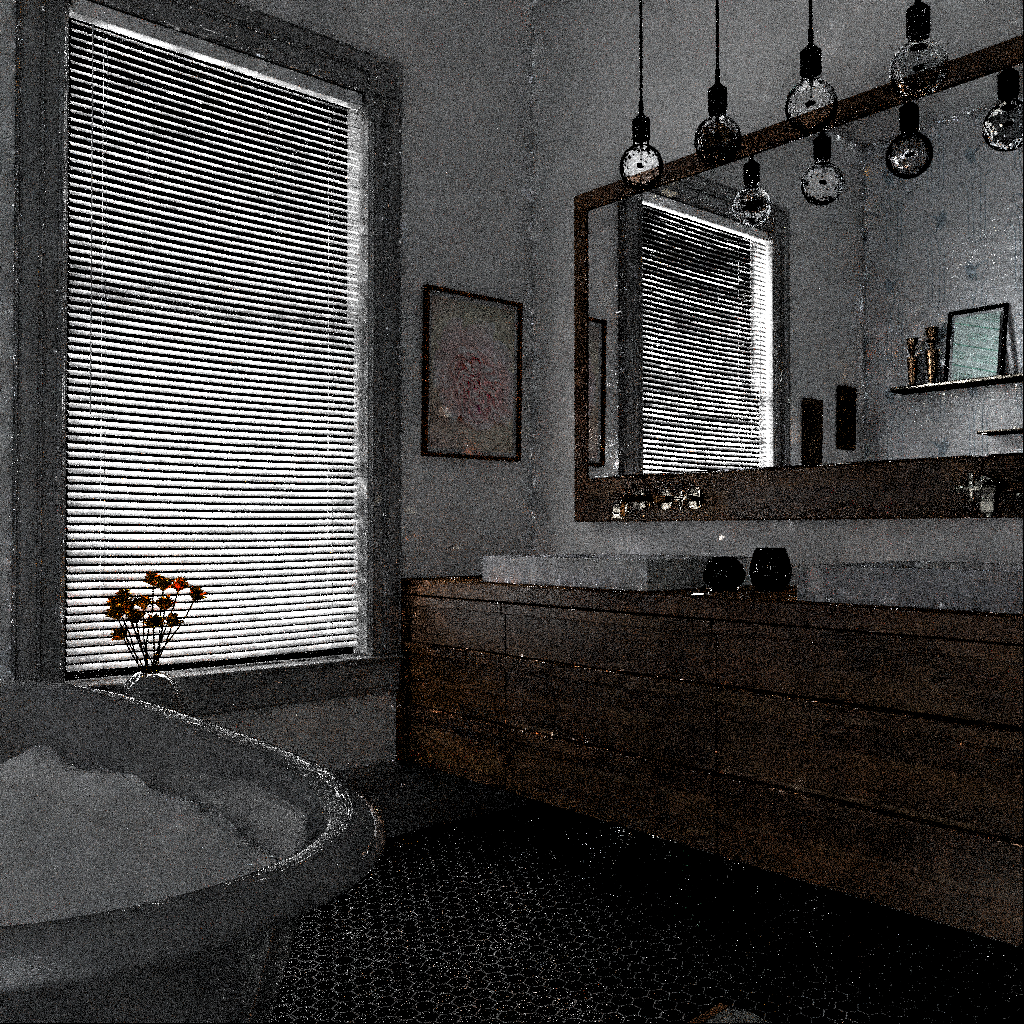} \\
\begin{tabular}{lll}
\includegraphics[width=0.1\textwidth]{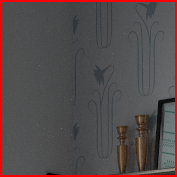} & \includegraphics[width=0.1\textwidth]{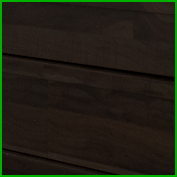} & \includegraphics[width=0.1\textwidth]{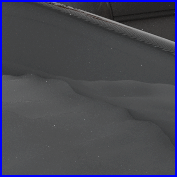}
\end{tabular} &
\begin{tabular}{lll}
\includegraphics[width=0.1\textwidth]{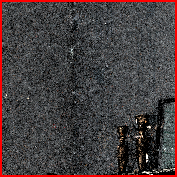} & \includegraphics[width=0.1\textwidth]{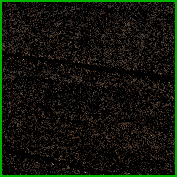} & \includegraphics[width=0.1\textwidth]{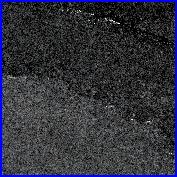}
\end{tabular} &
\begin{tabular}{lll}
\includegraphics[width=0.1\textwidth]{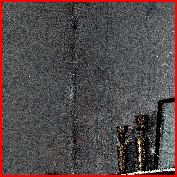} & \includegraphics[width=0.1\textwidth]{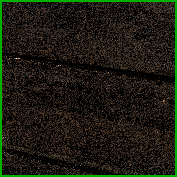} & \includegraphics[width=0.1\textwidth]{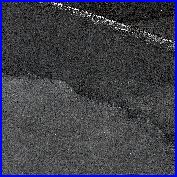}
\end{tabular} \\
Reference & MLT & Ours
\end{tabular}
\end{center}
\caption{Our approach splits path space into a discrete set of partitions, each of which can be integrated by a separate estimator. As these are now integrating over sparser spaces, we propose a guided image plane sampling approach based on an analysis of the acceptance probability for image plane perturbations and accelerated by denoising the information used to create the partitions. This image shows the \emph{bathroom} scene showing variance is reduced using our approach (on the right) and Metropolis Light Transport (in the middle) computed at the same number of samples.}
\label{fig:teaser}
\end{figure}

\section{Introduction}

Modern rendering algorithms rely on well established approaches for integrating over light paths. These typically find efficient ways of generating paths which connect light sources to the camera via a series of interactions with a scene. To render scenes containing complicated light transport efficiently, various methods have been proposed, such as Bidirectional Path Tracing \cite{lafortune1993bi, veach1995bidirectional}, ReSTIR approaches \cite{lin2022generalized, kettunen2023conditional} and Markov Chain Monte Carlo \cite{veach1997metropolis, kelemen2002simple, hachisuka2014multiplexed}. Many of the concepts in these works overlap, for instance exploiting the correlation between nearby paths in path or image space.

This work develops an approach for partitioning path space to further exploit these correlations, whre each of these partitions can be integrated by a separate estimator. This can be more efficient than purely integrating over path space as a whole, and in this work we focus on integrating using MCMC methods as these form a useful starting point for developing traditional Monte Carlo estimators.

However, these partitions of path space are sparser than path space, and this poses a challenge to generate valid paths within each subspace. This means effective perturbation strategies need to be aware of the partitioned space, and constrain proposal distributions to this space. Motivated by the success of other image space and low dimensional integration approaches, we propose utilizing information in image space to guide perturbations.

We achieve the partitioning by using a Monte Carlo path tracing pre-pass to estimate partitions of path space, then also denoising the contributions of the paths within each partition to build the image plane guidance distribution. To summarise, our contributions are as follows:

\begin{itemize}
    \item A principled approach to partition path space into subspaces, each of which can be integrated by a separate estimator
    \item An image plane path guidance distribution to generate proposals which explore this partitioned path space, and propose a method to use sparse image plane information to build this proposal distribution.
    \item Results for our approach applied to path space MLT algorithms showing improvements in image quality.
\end{itemize}

\section{Related Work}
\label{sec:relatedwork}

Our work focuses on partitioning path space and guiding perturbations so we briefly review related work on MCMC algorithms and path guiding.

MCMC was first applied to rendering in Metropolis Light Transport by\cite{veach1997metropolis} who formulated the now ubiquitously used path space formalism of light transport. This applied Metropolis sampling \cite{metropolis1953equation, hastings1970monte} over a space of all possible paths, and achieved a substantial improvement in scenes containing hard to sample paths over conventional techniques such as path tracing \cite{kajiya1986rendering} or bidirectional path tracing \cite{lafortune1993bi, veach1995bidirectional}. The main advantage of this approach was the local exploration of space around an existing path, allowing for substantially more non-zero contribution paths to be created at the same computational cost.

Path space MCMC has been further developed in several directions, for example \cite{pauly2000metropolis} extended MLT to participating media \cite{pauly2000metropolis}. \cite{jakob2012manifold} proposed a perturbation based on manifold walks which efficiently connects path vertices through specular or near-specular interactions. \cite{otsu2018geometry} proposed a perturbation strategy which adapted the perturbation size based on the local geometry, meaning that perturbations adapted well to high frequency geometry. Half vector space was used in \cite{kaplanyan2014natural} to constrain perturbations to a path and had the elegant property of canceling out most geometry terms in the acceptance probability. \cite{bashford2021ensemble} proposed several perturbations based on using an ensemble of paths to guide sampling of an individual path. \cite{manzi2014improved} proposed a series of improvements which led to a decrease in variance in Metropolis sampling for gradient domain rendering, and used image space exploration to find proposed paths, although in the context of finding pixel shifts rather than a general perturbation strategy. Other approaches such as Energy Redistribution Path Tracing \cite{cline2005energy}, Multiple-Try MCMC  \cite{segovia2007metropolis, nimier2019mitsuba}, and Delayed Rejection \cite{Rioux-Lavoie:2020:DRMLT} all improve MCMC methods by proposing the use of multiple short chains or multiple attempts at creating a proposal distribution respectively.

\cite{kelemen2002simple} proposed Primary Sample Space MLT (PSSMLT), an alternative approach to path space MCMC which operates on the random numbers used to generate paths. This formulation had the advantage that it is substantially simpler than path space methods, yet could still explore local regions of path space. \cite{hachisuka2014multiplexed} operates in a space of multiple paths, each of a different length, and selects the number of vertices to perturb from the light and camera. Multiple Importance Sampling is used to weight the path contributions of the selected combination of light and eye path. This splitting path space into paths of the same length is close to our work, and we discuss this more in Section \ref{sec:PartitioningPractical}.

Several other works have extended PSSMLT. Work such as \cite{li2015anisotropic} and \cite{luan2020langevin} used local path gradient information to generate anisotropic proposal distributions which are very effective at adapting proposals in path space. These are complementary to our approach which uses \emph{global} information in image space to construct a proposal distribution. \cite{sawhney2022decorrelating} recently proposed the use of PSS perturbations to decorrelate contributions in ReSTIR algorithms \cite{lin2022generalized}. Path space MLT and PSSMLT were combined into a single space by several works \cite{otsu2017fusing, pantaleoni2017charted,bitterli2018reversible}, and again our approach is complementary to this work.

Path guiding uses information gathered during rendering or via a pre-pass to build distributions which match the integrand in the Rendering Equation \cite{kajiya1986rendering} better than purely sampling the BSDF and cosine term. This encompasses a large amount of work, from guided sampling of whole paths \cite{reibold2018selective}, use of incident radiance stored on basis functions in world space \cite{jensen1995importance,hey2002importance,bashford2012significance,vorba2014online, herholz2016product,ruppert2020robust,diolatzis2020practical}, 5D Trees \cite{lafortune19955d, muller2017practical} to selective sampling for specific transport phenomena \cite{fan2023manifold, yu2023neural}. Our work generates samples in image space, and the closest guiding approach is \cite{cline2008table} who store information about recent sampling decisions in image space and use this to guide future sampling. This however is not directly applicable to our work as this uses a small cache of sampling decisions in image space, and is based on traditional Monte Carlo sampling rather than MCMC.

\begin{figure*}[htp]
    \includegraphics[width=\textwidth]{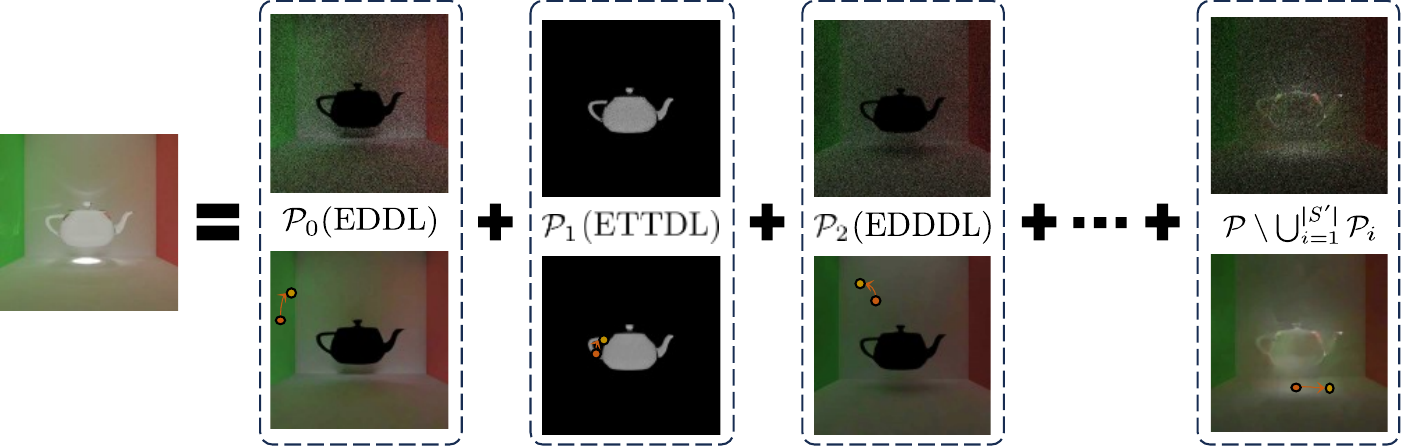}
    \caption{Our approach partitions the whole of path space (the left image) into a series of partitions, discussed in Section \ref{sec:partitioning}, each corresponding to a different subset of path space. We propose splitting on the interaction types, as illustrated in the boxes on the right, with the rightmost box illustrating the contribution from the complementary partition (Equation \ref{eq:fullpartitions}). Each of these partitions are constructed by performing a Monte Carlo sampling pre-pass to find the contribution of each found interaction type (top images, and discussed in Section \ref{sec:PartitioningPractical}). These contributions are denoised and are used to guide MCMC sampling (proposals illustrated in the denoised bottom images), see Section \ref{sec:GIS}.}
    \label{fig:partitions}
\end{figure*}

\section{Background}

To begin, we start by introducing the path integral form of light transport upon which most rendering algorithms are based. This was described by \cite{veach1997metropolis, hachisuka2014multiplexed} as:

\begin{equation}
I_{j} = \int_{\mathcal{P}} h_{j}(\overline{x}) f(\overline{x}) d_{\mu}(\overline{x}),
\label{eq:PathIntegral}
\end{equation}

\noindent where the value of the $j$'th pixel $I_{j}$ in an image is given by integrating the contribution of a light path $f(\overline{x})$ weighted by a filter at the pixel $h_{j}(\overline{x})$. A path is defined as a series of path vertices $x_{0}..x_{M}$, and integration is typically performed with respect to the product area measure $\mu$. The contribution of a path is then defined as the product of terms associated with interactions with a scene (note for notational convenience we index path vertices starting at the camera):

\begin{align}
f(\overline{x}) =& G(x_{0} \leftrightarrow x_{1}) \nonumber \\& \left[\prod^{M-1}_{k=1}fr(x_{k-1}\rightarrow x_{k}\rightarrow x_{k+1})G(x_{k} \leftrightarrow x_{k+1}) \right] Le(x_{M}),
\label{eq:PathContrib}
\end{align}

\noindent where $fr(x_{k-1}\rightarrow x_{k}\rightarrow x_{k+1})$ is the BSDF and $G(x_{k} \leftrightarrow x_{k+1}) = \frac{\cos(\theta)\cos(\theta')}{||x_{k} - x_{k+1}||_{2}}V(x_{k} \leftrightarrow x_{k+1})$ is the Geometry Term where $\theta$ and $\theta'$ are the angles between the surface normals and the outgoing direction from the surface. $V(x_{k} \leftrightarrow x_{k+1})$ denotes visibility between two points.

This is typically integrated over the set of all paths $\mathcal{P}$ which is defined as the union of all path lengths $\mathcal{P} = \bigcup_{i=2}^{\infty}\mathcal{P}(i)$ where $\mathcal{P}(i)$ are all paths of length $i$ which connect the camera to the light source. This also indicates that paths can be used to integral all pixels in an image, but only paths for which $h_{j}(\overline{x}) > 0$ will contribute to the $j$'th pixel.

Equation \ref{eq:PathIntegral} can be solved with several numerical methods, but in this work we focus on MCMC approaches. Metropolis Sampling \cite{metropolis1953equation, hastings1970monte} starts from an initial state $\overline{x}$, and proposes a new tentative state $\overline{x'}$ by sampling from a proposal distribution $T(\overline{x} \rightarrow \overline{x'})$ and the state is updated according to an acceptance probability:

\begin{equation}
a(\overline{x} \rightarrow \overline{x'}) = min\left(1, \frac{f^{*}(\overline{x'})T(\overline{x'} \rightarrow \overline{x})}{f^{*}(\overline{x})T(\overline{x} \rightarrow \overline{x'})}\right),
\label{eq:AcceptanceProbabilityOriginal}
\end{equation}

\noindent where $f^{*}(\overline{x})$ is the scalar contribution function which maps RGB or spectral radiance to a scalar. This has been shown to generate states which are distributed proportional to $f^{*}$ while allowing more flexibility than a traditional Monte Carlo estimator to explore state space. \cite{veach1997metropolis} used this to solve Equation \ref{eq:PathIntegral} via the following estimator:

\begin{equation}
I_{j} \approx \frac{b}{N}\sum_{k=1}^{N}\frac{h_{j}(\overline{x}_{k})f(\overline{x}_{k})}{f^{*}(\overline{x}_{k})},
\label{eq:MCMCEstimator}
\end{equation}

\noindent where $b$ is a normalizing constant used to appropriately scale the histogram estimated by MCMC algorithms. $b$ can be estimated via a separate Monte Carlo estimator: $b = \int_{\mathcal{P}} f^{*}(\overline{x}) d_{\mu}(\overline{x})$.

\section{Path Space Partitioning}
\label{sec:partitioning}

Path space was defined in the previous sections as the union of paths of different lengths. While this is a general form of writing this, an alternative, and sometimes advantageous formulation is to partition path space into a set $S'$ of $K$ discrete subsets:

\begin{equation}
    S' = \{\mathcal{P}_{0}, \mathcal{P}_{1}, ... , \mathcal{P}_{K}\}.
    \label{eq:partitions}
\end{equation}

To guarantee full coverage of path space, we need to augment this set with the remaining paths in path space not covered by this set (we refer to this as the complementary partition):

\begin{equation}
    S = \{S', \mathcal{P} \setminus \bigcup_{i=1}^{|S'|}\mathcal{P}_{i}\}.
    \label{eq:fullpartitions}
\end{equation}

This means that for the $i$'th partition of path space, Equation \ref{eq:PathIntegral} can be written as:

\begin{equation}
I_{i,j} = \int_{\mathcal{P}_{i}} h_{j}(\overline{x}) f(\overline{x}) d_{\mu}(\overline{x}),
\label{eq:PathIntegralPartitioned}
\end{equation}

\noindent where each path $\overline{x} \in \mathcal{P}_{i}$. A MCMC estimator for Equation \ref{eq:PathIntegralPartitioned} can be written similar to Equation \ref{eq:MCMCEstimator}:

\begin{equation}
I_{i,j} \approx \frac{b_{i}}{N}\sum_{k=1}^{N}\frac{h_{j}(\overline{x}_{k})f(\overline{x}_{k})}{f^{*}(\overline{x}_{k})},
\label{eq:MCMCEstimatorPartitioned}
\end{equation}

\noindent where $b_{i}$ is defined as before but with respect to the partitioned path space $b_{i} = \int_{\mathcal{P}_{i}} f^{*}(\overline{x}) d_{\mu}(\overline{x})$. The final value of the pixel is clearly then the sum over integrals over each partition:

\begin{equation}
I_{j} = \sum^{|S|}_{i = 1} I_{i,j}.
\label{eq:PathIntegralPartitioned}
\end{equation}

Equation \ref{eq:MCMCEstimatorPartitioned} can also be written as a Monte Carlo estimator where each partition is selected with probability $P(i)$:

\begin{equation}
I_{j} \approx \frac{1}{N}\sum_{k=1}^{N}\frac{b_{i}h_{j}(\overline{x}_{k})f(\overline{x}_{k})}{P(i)f^{*}(\overline{x}_{k})}.
\label{eq:MCMCEstimatorPartitioned}
\end{equation}

While this might initially seem to add unnecessary complication, there is an advantage to this formulation. MCMC algorithms transition between states proportional to their contribution. This means that the sampling algorithm will spend more time in states with a higher scalar contribution function than those with a lower contribution. While this is to be expected, in applications such as rendering this is suboptimal as scenes in which MCMC algorithms are effective often have widely varying values on the image plane. An example of this is path spaces which include specular interactions such as caustics which often lead to small regions of the scene having values which are orders of magnitude larger than the majority of the scene. As a result, the chain spends orders of magnitude more time in these regions, which comes at the cost of more variance in darker regions in an image.

Therefore, if we can partition path space into partitions which have similar contributions, then rather than chains spending a significant amount of time in brighter regions, each chain will explore its own reduced space. Chains in bright regions, with associated higher normalizing constant $b_{i}$, will only explore bright regions, and chains associated with darker regions will be able to expend more computation in these regions, thereby reducing variance in these regions.

\subsection{1D Example}

\begin{figure}[tp]
\centering
\setlength{\tabcolsep}{0pt}
\begin{tabular}{cc}
 \includegraphics[width=0.5\linewidth]{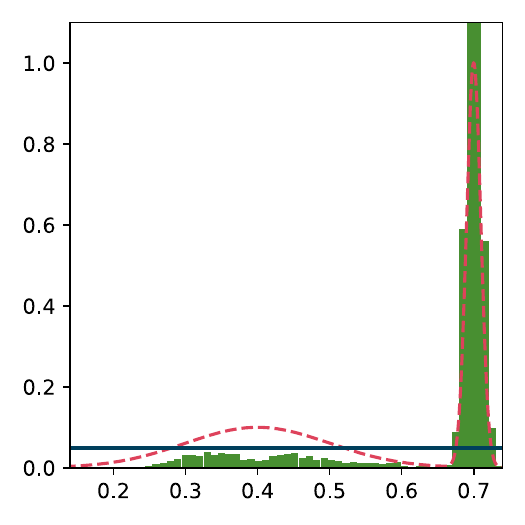} &
     \includegraphics[width=0.5\linewidth]{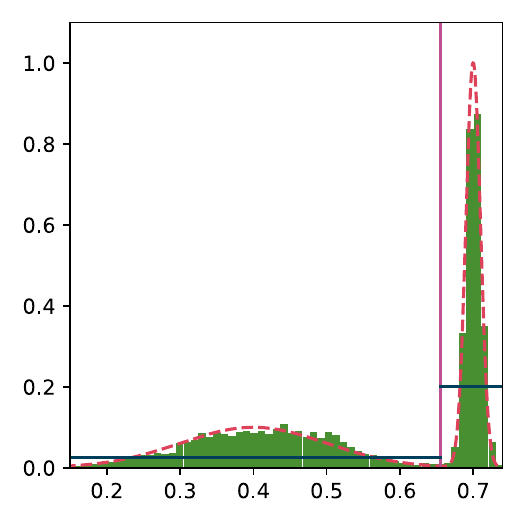} \\
a) Without partitions  & b) With Partitions
\end{tabular}
\caption{Example in 1D of using partitions. a) and b) show MCMC integration with and without partitioning (the partition is shown by the vertical line in b)). Using partitioning decreases variance significantly by allowing two chains to explore the lower and higher contributions separately, each of which uses a different normalizing constant (the horizontal line).}
\label{fig:1DExample}
\end{figure}

As an example of this, Figure \ref{fig:1DExample} shows a 1D example of using partitioning. \ref{fig:1DExample} a) and \ref{fig:1DExample} b) show a function (red dashed line) being integrated using standard MCMC sampling without (a)) and with (b)) partitioning. Both use the same gaussian proposal distribution, but b) uses two chains, one for the left side of the partition (shown by the vertical line) and one for the right. Using two chains in this example allows one chain to explore the high contribution region on the right, and one to explore the left, low contribution region. This means that both regions are sampled adequately given the low sample count used in these figures. The horizontal lines indicate the value of the normalizing constant in each region. In contrast not using partitions leads to higher variance in both regions as the chain has explored the higher contribution region excessively at this low sample count. Note that both a) and b) converge to the correct value, but b) has 89 times less variance in this example.

\begin{figure*}[htp]
\centering
     \begin{subfigure}[b]{0.24\textwidth}
         \centering
         \includegraphics[width=\textwidth]{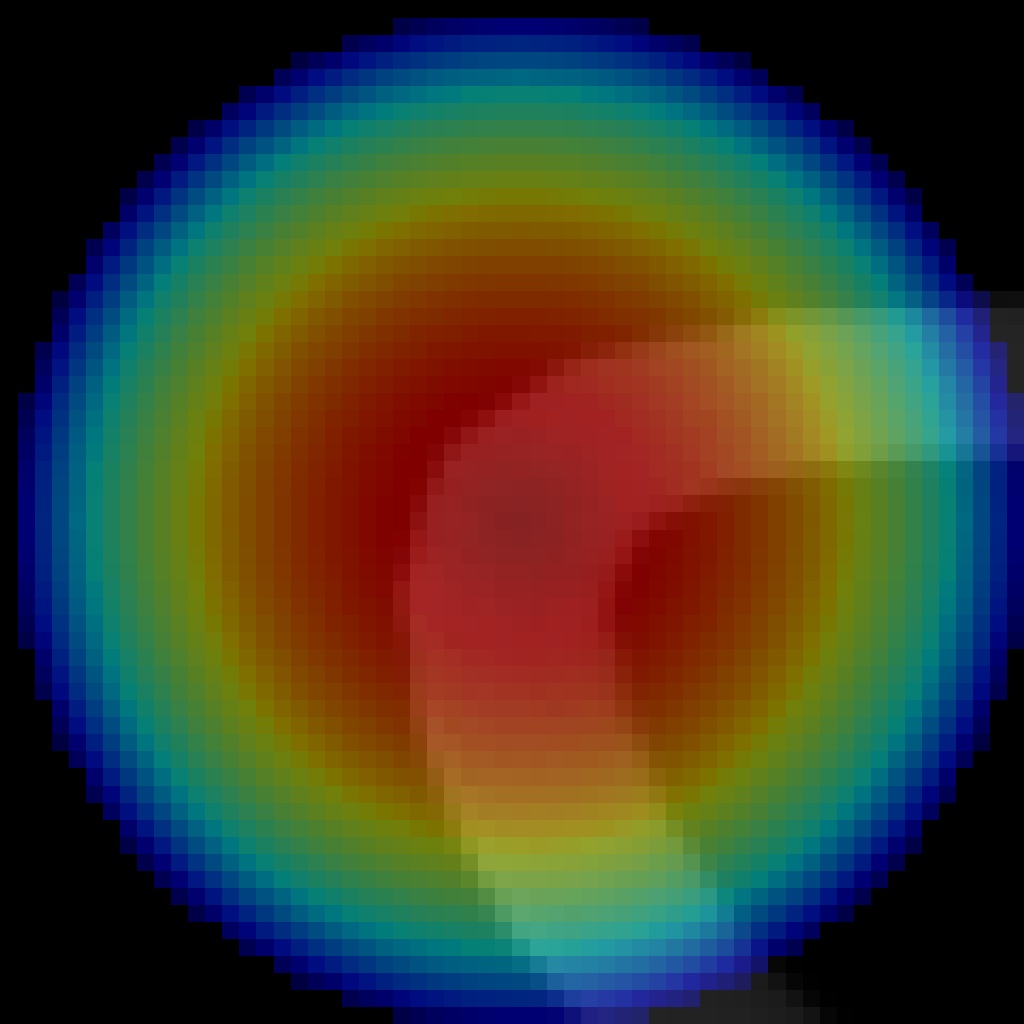}
         \caption{Isotropic}
         \label{fig:kiso}
     \end{subfigure}
     \begin{subfigure}[b]{0.24\textwidth}
         \centering
         \includegraphics[width=\textwidth]{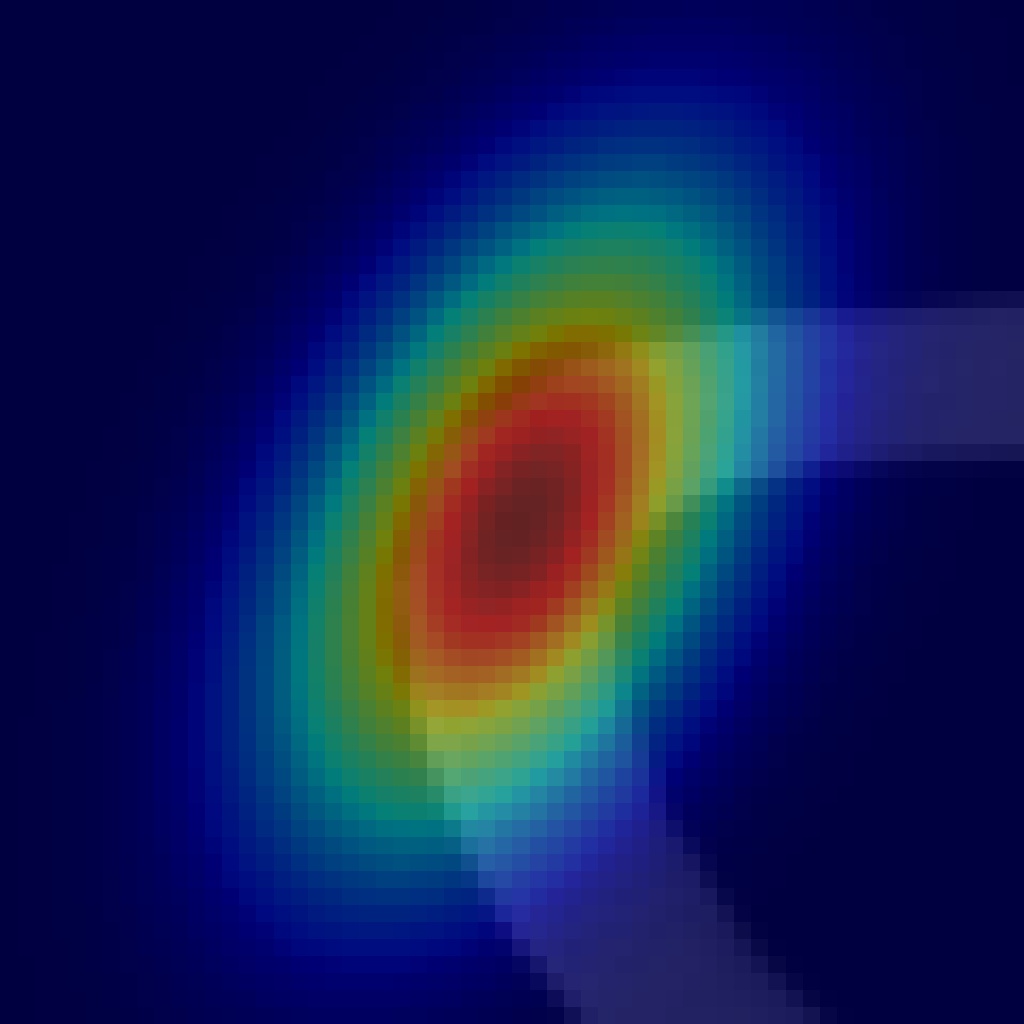}
         \caption{Anisotropic}
         \label{fig:kaiso}
     \end{subfigure}
     \begin{subfigure}[b]{0.24\textwidth}
         \centering
         \includegraphics[width=\textwidth]{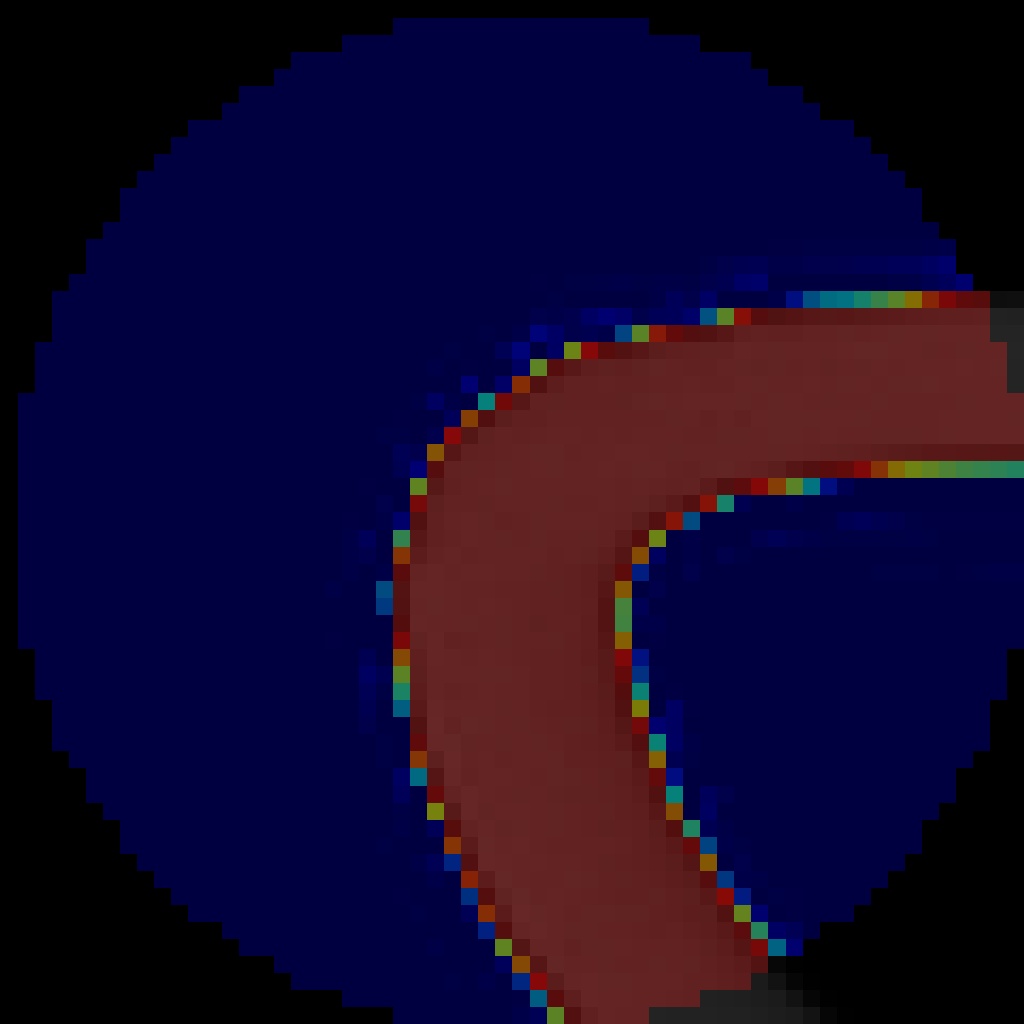}
         \caption{$\emph{Full } Y'$}
         \label{fig:kexact}
     \end{subfigure}
     \begin{subfigure}[b]{0.24\textwidth}
         \centering
         \includegraphics[width=\textwidth]{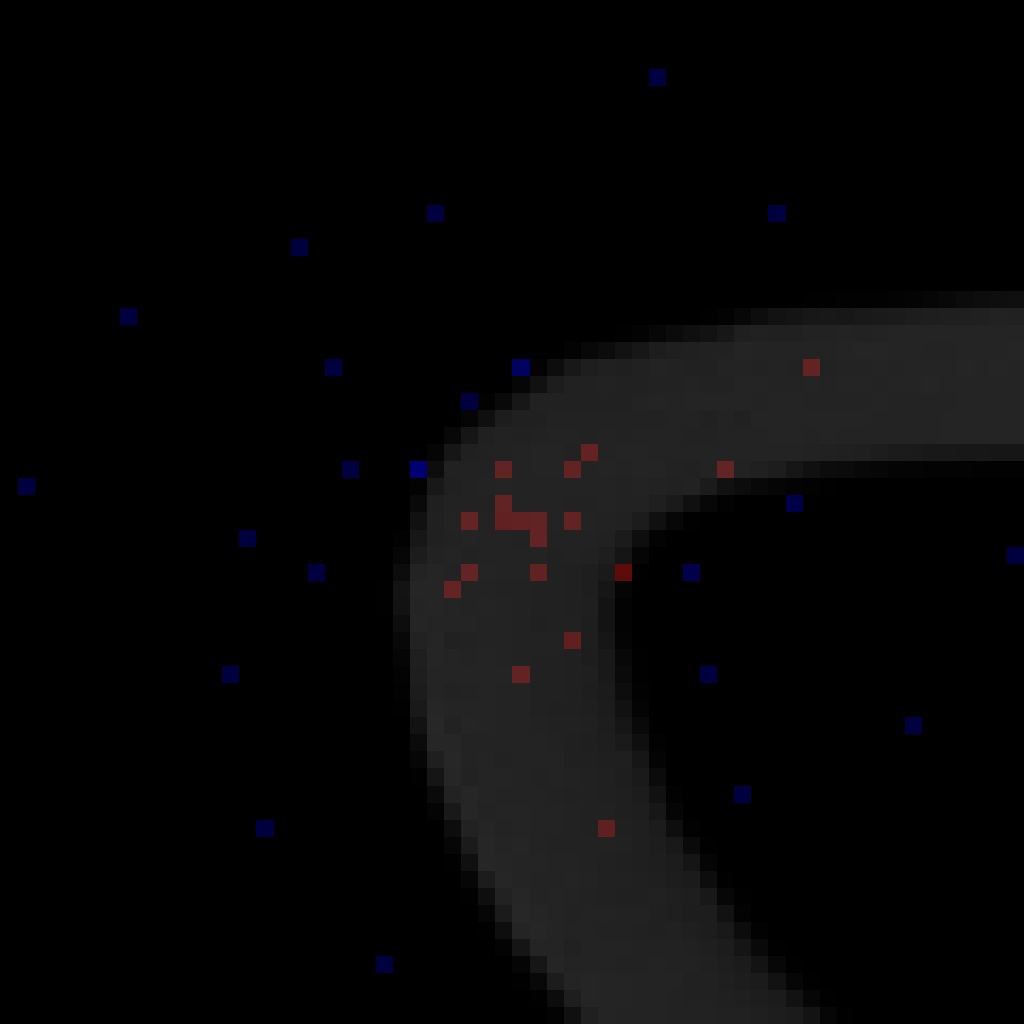}
         \caption{$\emph{Sparse } Y'$}
         \label{fig:ksparse}
     \end{subfigure}
\caption{Illustration of different image space proposal distributions for a region in a partition from Figure \ref{fig:partitions}, red regions indicate higher and blue lower values of the proposal distribution. Isotropic approaches (\ref{fig:kiso}) do not use any information and can propose paths that are likely to be rejected. Anisotropic proposals (\ref{fig:kaiso}) adapt better to the distribution of radiance but are still limited to a parametric distribution. Figure \ref{fig:kexact} shows our approach which adapts \emph{per-pixel} to the estimated lighting based on the denoised estimate of lighting per partition $D_{i}$, and Figure \ref{fig:ksparse} shows our sparse approximation.}
\label{fig:proposalkernels}
\end{figure*}

\subsection{Practical Considerations}
\label{sec:PartitioningPractical}

Now we have a formalism to partition path space, this leaves three questions: how many partitions ($K$) should be used, how should path space be partitioned, and with what probability should each partition be sampled ($P(i)$)?

There has been some previous work which has considered this problem, i.e. \cite{hachisuka2014multiplexed}, \cite{bashford2021ensemble} and \cite{lin2022generalized}. \cite{hachisuka2014multiplexed} is the most similar as they explicitly integrate in a set of partitions of different path lengths, i.e.

\begin{equation}
    S_{h} = \{\mathcal{P}_{0}, \mathcal{P}_{1}, ... , \mathcal{P}_{K}\},
\end{equation}

\noindent where $\mathcal{P}_{i}$ denotes a path space with $i + 4$ vertices. Each of these partitions was sampled proportional to its contribution, and path lengths up to $K + 4$ could be computed. However, while this solution is unbiased and quite effective for these paths, this neglects the rest of path space from Equation \ref{eq:fullpartitions}, and faces the same issue that paths of the same length may explore regions of path space with contributions of significantly different orders of magnitude. For example, using Heckbert path notation \cite{heckbert1990adaptive}, a path of length 5 may include both low contribution $LDDDE$ paths, but very high contribution $LSSDE$ paths.

We propose a solution to the three questions. Like \cite{hachisuka2014multiplexed}, we use a Monte Carlo estimator as a pre-pass and trace a number of paths. Each subpath can be viewed as estimating a separate integral, i.e. estimating a value of a partition of path space, as the path is progressively constructed. Therefore this pre-pass gives us much of the information needed to partition path space if we can efficiently extract this information. At this point any information about each sub-path could be used, for example interaction types which classify vertices into specular or diffuse akin to Heckert notation, regions of the scene or image space explored by the sub path etc. We take the approach of storing paths corresponding to a unique set of interactions as specified by their interaction types.

Practically, we store a linked list of buffers $B$ corresponding to each interaction type (i.e. $LDDE$, $LSDE$, $LDDDE$), and accumulate the contribution of this sub path ($C(\overline{x}_{k})=\frac{f(\overline{x}_{k})}{p(\overline{x}_{k})}$, where $p(\overline{x})$ is the pdf of generating this path), the seed used to generate this sub-path. We store this as a linked list to minimize the memory requirements associated with the combinatorial explosion of interaction types in a complicated scene.

As each buffer defines a partition of path space, we can map some attributes from each buffer to a scalar and can calculate the $K$ most important partitions for use in Equation \ref{eq:partitions}. The mapping can use any heuristic function, however, we want to both consider the contribution of the paths in the partition as well as the area of the image plane over which the paths explore. The reason for this is that we want to explore large, low contribution regions as well as small high contribution regions. Therefore, we choose to consider the total contribution of half the paths $|B|/2$ (the other half is used for initialization of the MCMC algorithm) in the $i$'th partition ($\gamma(i) = \sum^{|B|/2}_{k=1} C(\overline{x}_{k})$ as this accounts for both the contributions of the paths and the area of the image covered by these paths.

We can then sort the buffers by $\gamma$ and choose the $K$ largest which fit within the memory requirements as discussed in Section \ref{sec:GIS}. The remaining paths are assigned to the complementary partition (see Equation \ref{eq:fullpartitions}). The partition is then defined by the path interaction types in the partition and the probability of choosing a partition is then $P(i) = \frac{Y(i)}{\sum^{K}_{k=1}\gamma(k)}$.

We now need to compute the normalization constant and initialize each MCMC chain in each partition. To estimate both, we split the initial samples used into two sets of size $|B|/2$. One set is used to estimate the partitions as discussed above, and the contributions from the other are used to compute the normalizing constant for the partition ($b_{i}$). Furthermore, an initial path is resampled from this other set, and we also run burn-in \cite{brooks2011handbook} for 1024 iterations to further reduce startup bias.

\section{Guided Image Sampling}
\label{sec:GIS}

Partitioning path space as described in the previous section leads to a discrete set of regions of path space that will be explored by a chain. However, this partitioning of path space leads to sparser regions of path space which contain valid contributions on the image plane. In image space, this often means that paths from one partition can only contribute within a small region, but paths from other partitions may contribute widely over the image plane. If the original proposal distributions defined in MLT or PSSMLT are used, these are not likely to be able to explore these spaces well. Motivated by this, we investigate guiding perturbations on the image plane such that both small, sparse regions can be explored effectively, while also having the ability to widely and efficiently explore the image plane.

We propose to solve this via guiding perturbations on the image plane. There are two main reasons for choosing the image plane for guidance rather than the whole of path space. Firstly, in Path Space MLT \cite{veach1997metropolis}, for most scenes the lens perturbation is typically responsible for the majority of the variance reduction. Secondly, several other algorithms in rendering exploit combining sparser estimates of suffix or light paths with a denser sampling of prefix or camera paths, such as Instance Radiosity \cite{keller1997instant} and final gathering for photon mapping \cite{jensen2001realistic} or conditional ReSTIR \cite{kettunen2023conditional}, which again shows that well converged results can be obtained by focusing sampling on the image space.

We next discuss guided sampling in path space and formulate practical guidance distributions for both spaces.

\subsection{Path Space Image Plane Guiding}
\label{sec:pathspaceimageplaneguiding}

To generate guided samples on the image plane that can still explore the local region around the current path requires a proposal distribution which is aware of the contributions of a path to the image plane in a local region. States will still be visited proportional to their contribution, but we aim to derive a proposal distribution that can increase the probability of moving between states based on their contribution rather than a fixed proposal distribution.

We start by considering the acceptance probability computation given in Equation \ref{eq:AcceptanceProbabilityOriginal}.  Perturbing a path in image space means that we are perturbing vertices successively until they can be connected to a fixed remainder of a path which starts at the $s$'th vertex from the camera, which has the contribution:

\begin{align}
    \alpha(\overline{x}) = &G(x_{s} \leftrightarrow x_{s + 1}) \nonumber \\ &\left[\prod^{M-1}_{k=s+1}fr(x_{k-1}\rightarrow x_{k}\rightarrow x_{k+1})G(x_{k} \leftrightarrow x_{k+1}) \right] Le(x_{M}).
\end{align}

Therefore, the prefix path we want to perturb has the following contribution:

\begin{align}
    S(\overline{x}) =& G(x_{0} \leftrightarrow x_{1}) \left[\prod^{s}_{k=1} fr(x_{k-1}\rightarrow x_{k}\rightarrow x_{k+1})G(x_{k} \leftrightarrow x_{k+1})\right] \nonumber \\& fr(x_{s - 1}\rightarrow x_{s}\rightarrow x_{s+1}). \label{eq:prefixcontrib}
\end{align}

The acceptance probability can now be written as:

\begin{equation}
a(\overline{x} \rightarrow \overline{x'}) = min\left(1, \frac{S^{*}(\overline{x'})}{S^{*}(\overline{x})}\cancel{\frac{\alpha^*(\overline{x})}{\alpha^*(\overline{x})}} \frac{T(\overline{x'} \rightarrow \overline{x})}{T(\overline{x} \rightarrow \overline{x'})}\right),
\label{eq:AcceptanceProbabilityupdated}
\end{equation}

Therefore, $T(\overline{x} \rightarrow \overline{x'}) \propto S^{*}(\overline{x'})$ then states would be generated exactly proportional to their contribution up to a normalizing constant. This is challenging to evaluate analytically in a potentially wide region around the current path in image space. However, if a set of candidate prefix paths $Y' = \{\overline{x'}_{0}, \overline{x'}_{1},..,\overline{x'}_{|Y'|}\}$, each with contribution $Q' = \{S^{*}(\overline{x'}_{0}), S^{*}(\overline{x'}_{1}), .., S^{*}(\overline{x'}_{|Q'|})$ can be generated, then a candidate path $\overline{x'}$ can be sampled from $Y'$ proportional to $Q'$, thereby closely approximating the ideal distribution in a local region.

As we are perturbing in image space, the logical domain on which to create the set of paths is over pixels in the local image space neighborhood of the current path $\overline{x}$. Therefore, all we need is to evaluate $S^{*}$ for the subset of pixels. However, even though $S(\overline{x})$ does not involve many terms, it still requires, at a minimum, evaluating two BSDFs and two geometry terms (also considering that each geometry term includes a visibility test). However, we can leverage the Monte Carlo pre-pass from Section \ref{sec:PartitioningPractical} to substantially reduce the computation requirements by making a series of approximations.

Firstly, we assume that the last BSDF evaluation will be similar for all paths in $Y'$, and set $fr(x_{s - 1}\rightarrow x_{s}\rightarrow x_{s+1}) = 1$. Secondly, as the BSDFs in a local region often do not vary much we approximate all other BSDF evaluations as a diffuse BRDF, i.e. $\frac{\rho_{j}}{\pi}$, where $\rho_{j}$ is the albedo of the first non-specular vertex visible through pixel $j$. Thirdly, we can precompute $G(x_{0} \leftrightarrow x_{1})$ for all pixels as $G_{j}$. Finally, we use an approximate value for the visibility term ($V'_{j}$ in the final Geometry Term, which allows us to avoid tracing any visibility rays. We discuss this in Section \ref{sec:approxvis}. 

\begin{figure}[tp]
    \centering
    \includegraphics[width=0.3\textwidth]{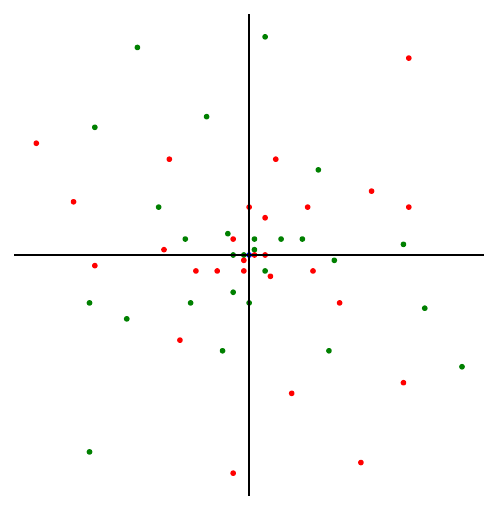}
    \captionsetup{margin={-10pt, -10pt}, justification=centering}
    \caption{Sparse offsets used to compute $Y'$. Green points are the arbitrarily chosen initial points and red are the inverse to guarantee reversibility.}
    \label{fig:sparsekernel}
\end{figure} 

We can then use this information to approximate $S(\overline{x})$ for the $j$'th pixel in the set $Y'$:

\begin{equation}
    S'(\overline{x}_{j}) \approx G_{j} \frac{\rho_{j}}{\pi} \frac{\cos(\theta_{j})\cos(\theta')}{||x_{s-1} - x_{s}||_{2}} V'_{j},
    \label{eq:prefixcontribapprox}
\end{equation}

\noindent where $\theta_{j}$ is the angle between a stored normal at the pixel $j$ and the outgoing direction. This leads to the final acceptance probability for the proposed guided image perturbation:

\begin{equation}
a(\overline{x} \rightarrow \overline{x'}) = min\left(1, \frac{S^{*}(\overline{x'})}{S^{*}(\overline{x})} \frac{S'^{*}(\overline{x})}{S'^{*}(\overline{x'})}\frac{\sum^{|Y'|}_{k=1}S'^{*}(\overline{x'}_{k})}{\sum^{|Y|}_{k=1}S'^{*}(\overline{x}_{k})}\right),
\label{eq:AcceptanceProbabilityproposed}
\end{equation}

The ratio of sums at the end of this expression comes from normalization constants that depend on both the set $Y'$ evaluated at pixels around $\overline{x}$, and also another set $Y$ which is evaluated at pixels around the proposed path $\overline{x'}$. This adds extra computation, so the size of $Y'$ must be chosen carefully to minimize computation cost.

We now discuss how to build $V'_{j}$, and which pixels to use for the set $Y'$.

\subsubsection{Building Image Space Approximations}
\label{sec:approxvis}

To build the image space approximations for $G_{j}$, $\rho_{j}$, and $V'_{j}$ we use the fact that we have stored the set of paths used to generate each partition, and that we can easily store other attributes such as image space position $X_{j}$, normal $n_{j}$, and albedo $\rho_{j}$ of the first non-specular vertex in a path in a GBuffer. $G_{j}$ and $\rho_{j}$ are then easy to precompute and store in an image space buffer, however, the visibility approximation is harder.

If we assume that the visibility will be similar in a local region of the image plane, then we can estimate an image space map of visibility for each of the pixels where paths from the partition can contribute. To achieve this, we leverage image space denoising (we use \cite{OpenImageDenoise}) of the paths from the Monte Carlo prepass which were used to estimate the partition. Note that any image space denoiser can be used for this, but instead of denoising all pixels in an image, the denoiser needs to focus on the smaller subset of pixels that have a non-zero contribution. This results in a denoised image for each partition $D_{i}$. We then directly leverage this to compute the visibility approximation:

\begin{equation}
    V'_{j} = \begin{cases}
        1 &\text{if } D_{i,j} > \epsilon\\
        \epsilon & \text{otherwise}
    \end{cases}
\end{equation}

\subsubsection{Choosing $Y'$}
\label{sec:sparsekernel}

We also have substantial freedom to choose the pixels which will form the elements of $Y'$ given one condition: the choice of elements of $Y'$ must preserve detailed balance, specifically the reversibility condition. Another way of saying this is that $T(\overline{x'} \rightarrow \overline{x})$ must be greater than zero for any proposed state $\overline{x'}$. An obvious approach could be to use all pixels within a fixed radius $R$ around and including the current state $\overline{x}$, referred to as $\emph{Full } Y'$. This guarantees that there is a non-zero density for reaching any pixel from any other pixel within $R$. We illustrate this in Figure \ref{fig:proposalkernels} which shows the difference between using isotropic proposal distributions, anisotropic proposals, and our approach.

This approach leads to a proposal distribution which adapts to the estimate of the illumination within a local region. However, computing the set of contributions $Q'$ is computationally expensive if all pixels within the region $R$ are considered. This cost can be substantially reduced by only considering a sparse approximation to $\emph{Full } Y'$ which obeys the above condition.

We describe the set of shifts in image space, measured in pixels, required to construct $Y'$ as a set of 2D offsets from the image plane coordinates of the current path:

\begin{equation}
\Delta = \{(\delta.x, \delta.y)_{0}, (\delta.x, \delta.y)_{1}, .., (\delta.x, \delta.y)_{|Y'|}\}.
\end{equation}

The first $\frac{|Y'|-1}{2}$ offsets in this set can be chosen arbitrarily, for example from a uniform sampling of a disk of radius $R$. The middle element needs to be $(0, 0)$, and the remaining elements of $\Delta$ must be set as the inverse of the first $\frac{|Y'|-1}{2}$ elements, e.g. $(\delta.x, \delta.y)_{\frac{|Y'|-1}{2} + k + 1} = (-\delta.x, -\delta.y)_{k}, \forall k \leq \frac{|Y'|-1}{2}$. This inversion guarantees that the initial state can be reached from any proposed state. We find it advantageous to propose more states closer to the current path, so we create the initial $\frac{|Y'|-1}{2}$ offsets using a low discrepancy sequence which generates points $(\zeta.x, \zeta.y)$ which we mapped to the disk using a non-uniformity preserving mapping to polar coordinates $(r, \Theta) = (\zeta.x^{2}, 2\pi \zeta.y)$ which are mapped to pixel shifts in $\Delta$. This is illustrated in Figure \ref{fig:sparsekernel} which shows the initial $\frac{|Y'|-1}{2}$ points in green, the center in blue, and the inverse set as red.

\section{Results}

\begin{figure*}[htp]
\setlength\tabcolsep{0pt}
\renewcommand{\arraystretch}{0}
\begin{center}
\begin{tabular}{cccccc}
Reference & Ref & MLT & GAMLT & EMLT & Ours\\
\multirow{3}{*}[0.15\textwidth]{\includegraphics[height=0.45\textwidth,keepaspectratio]{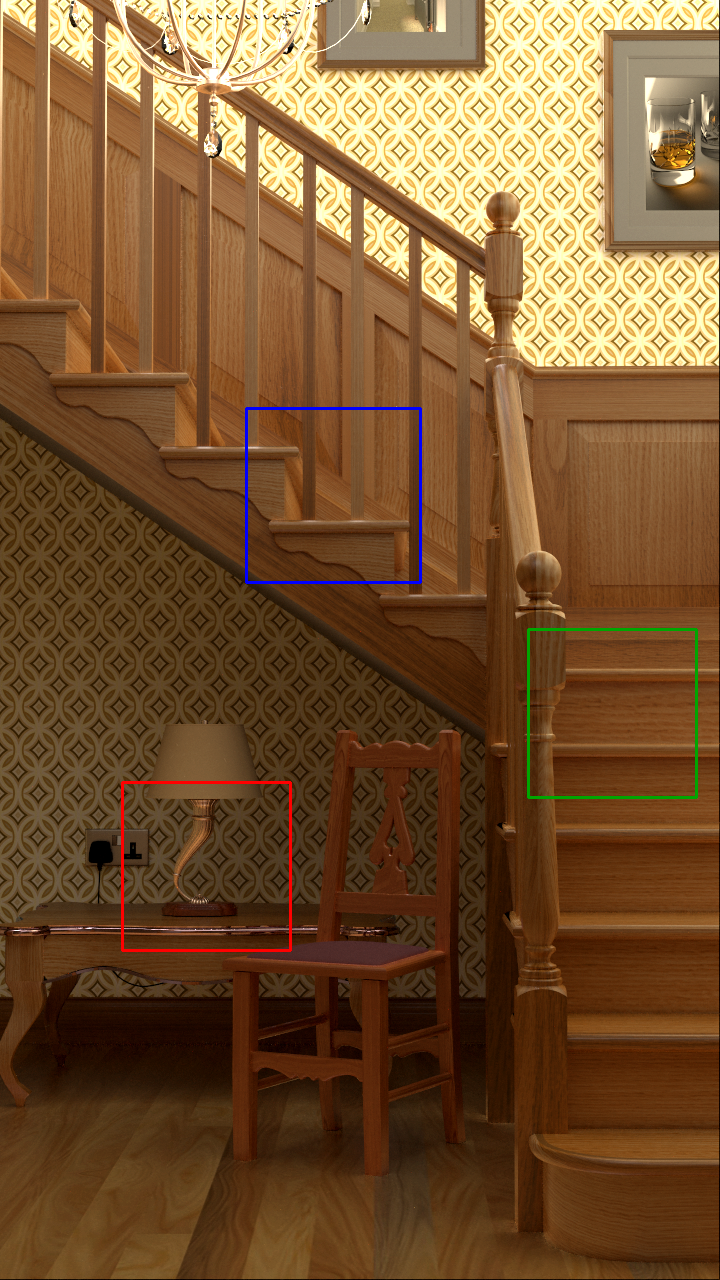}} & \includegraphics[width=0.15\textwidth]{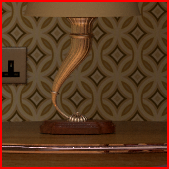} &
\includegraphics[width=0.15\textwidth]{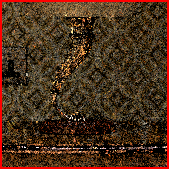} &
\includegraphics[width=0.15\textwidth]{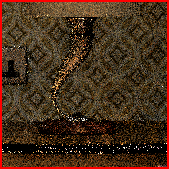} &
\includegraphics[width=0.15\textwidth]{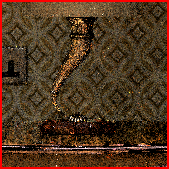} &
\includegraphics[width=0.15\textwidth]{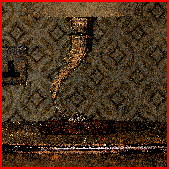} \\
& \includegraphics[width=0.15\textwidth]{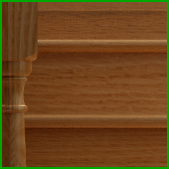} &
\includegraphics[width=0.15\textwidth]{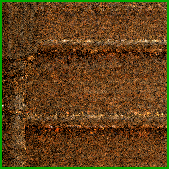} &
\includegraphics[width=0.15\textwidth]{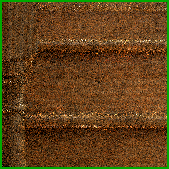} &
\includegraphics[width=0.15\textwidth]{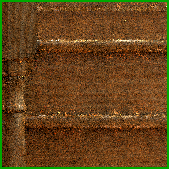} &
\includegraphics[width=0.15\textwidth]{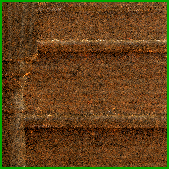} \\
& \includegraphics[width=0.15\textwidth]{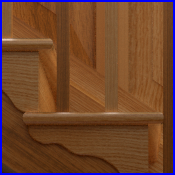} &
\includegraphics[width=0.15\textwidth]{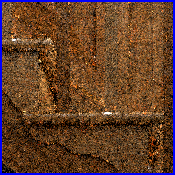} &
\includegraphics[width=0.15\textwidth]{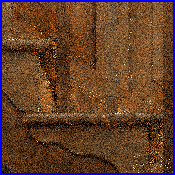} &
\includegraphics[width=0.15\textwidth]{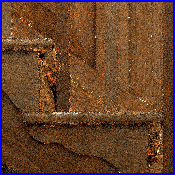} &
\includegraphics[width=0.15\textwidth]{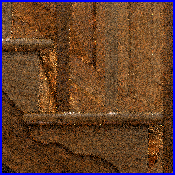}
\end{tabular}
\end{center}
\caption{Results for the \emph{staircase} scene comparing MLT, Geometry Aware MLT (GAMLT) with extensions proposed in \cite{bashford2021ensemble}, Ensemble Metropolis Light Transport (EMLT) and Ours.
}
\label{Fig:ResultStaircase}
\end{figure*}

We show results for our approach when applied to path space MLT \cite{veach1997metropolis}. We chose this algorithm for our baseline comparisons as it is the main path space MCMC integration algorithm, and as noted in Section \ref{sec:relatedwork}, our method is complementary to most approaches which extend these works. We used a CPU prototype implementation of all algorithms, and generated results on a laptop with a i9-11980 CPU and 32GB RAM. All methods use the same number of samples for initialization (computation of the normalization constants and resampling an initial path), but ours has an extra computational overhead of creating the denoised images, although this took a maximum of two seconds for all our results. We used the Intel Open Image Denoiser \cite{OpenImageDenoise} to denoise our images; and although this is not trained on denoising partitioned path space contributions on the image space, we found that it performed well. Our method has a linear memory cost in terms of the number of partitions, consisting of a tiny amount of extra memory to store per-partition chain information, and a larger resolution dependent buffer for use when evaluating $V'_{j}$ in Equation \ref{eq:prefixcontribapprox} This added around 80MB memory usage for all scenes in our results. We use 10 partitions plus the complementary partition in all our results as we found this allows the important majority of lighting to be found for all scenes, and use $|Y'| = 128$, as we found this balanced variance reduction with computational cost (our unoptimized implementation has a 19\% overhead compared to MLT)

All scenes were rendered on average at 32 mutations per pixel. Figure \ref{fig:teaser} shows the \emph{bathroom} scene, where our method has an RMSE of 0.4462 and MLT 0.5788, an 29\% improvement. Figure \ref{Fig:ResultStaircase} compares our approach (RMSE 0.00816) with MLT (RMSE 0.01156), Ensemble Metropolis Light Transport (EMLT 0.00742) \cite{bashford2021ensemble} and Geometry Aware MLT (GAMLT 0.00927) \cite{otsu2018geometry} with the extensions proposed in \cite{bashford2021ensemble}. While EMLT slightly outperforms our method, we remain competitive both in terms of objective metrics and visually as can be seen in the figure.

\begin{figure*}[htp]
\centering
\setlength\tabcolsep{0pt}
\renewcommand{\arraystretch}{0}
\begin{tabular}{ccc}
\includegraphics[width=0.333\textwidth]{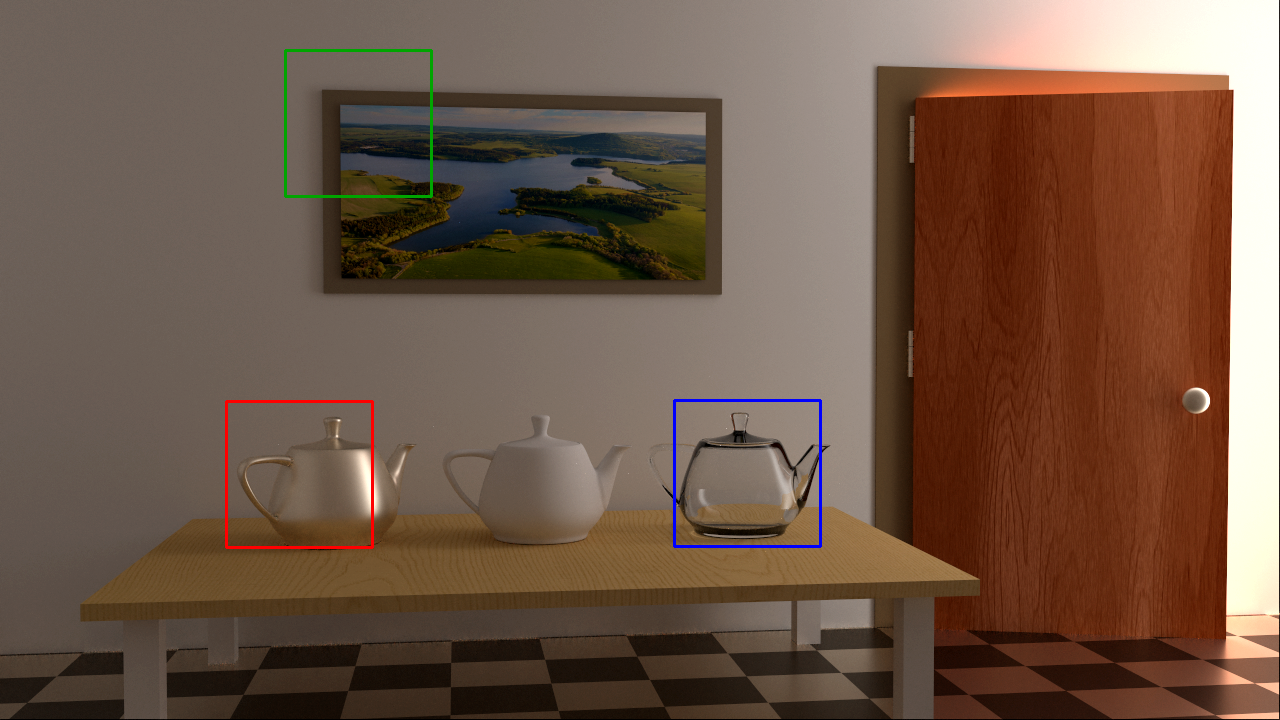} & \includegraphics[width=0.333\textwidth]{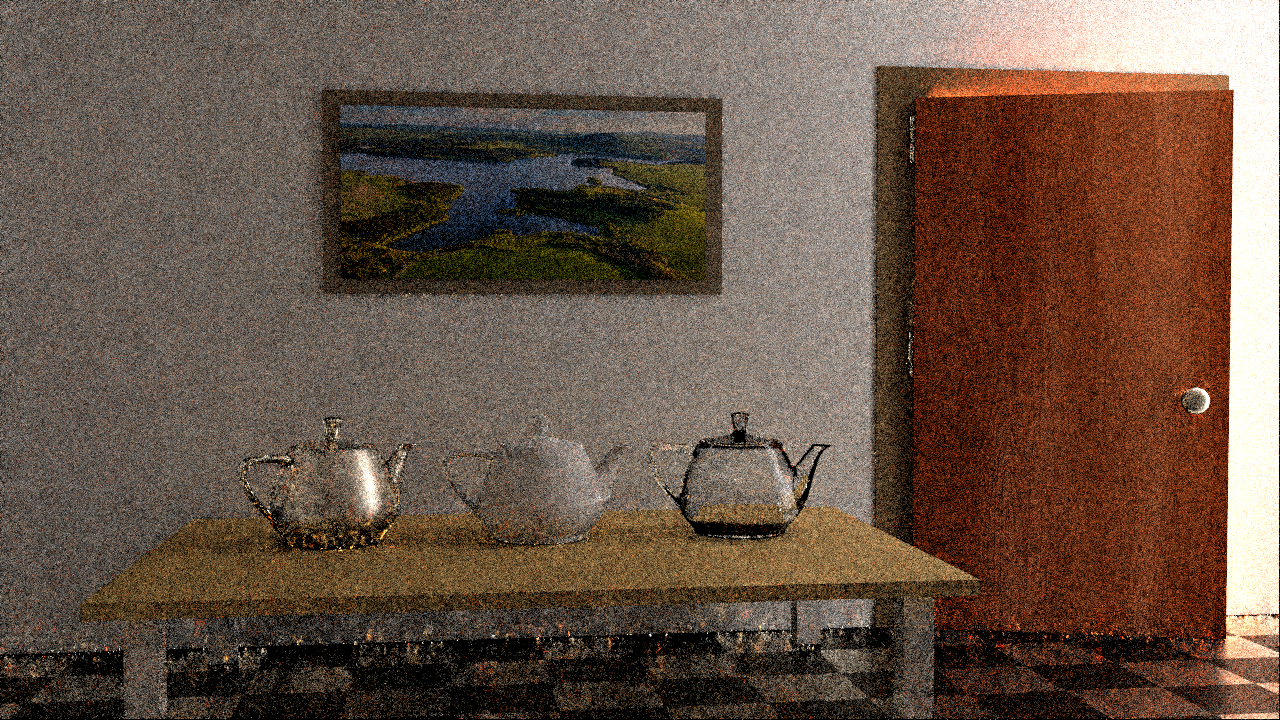} & \includegraphics[width=0.333\textwidth]{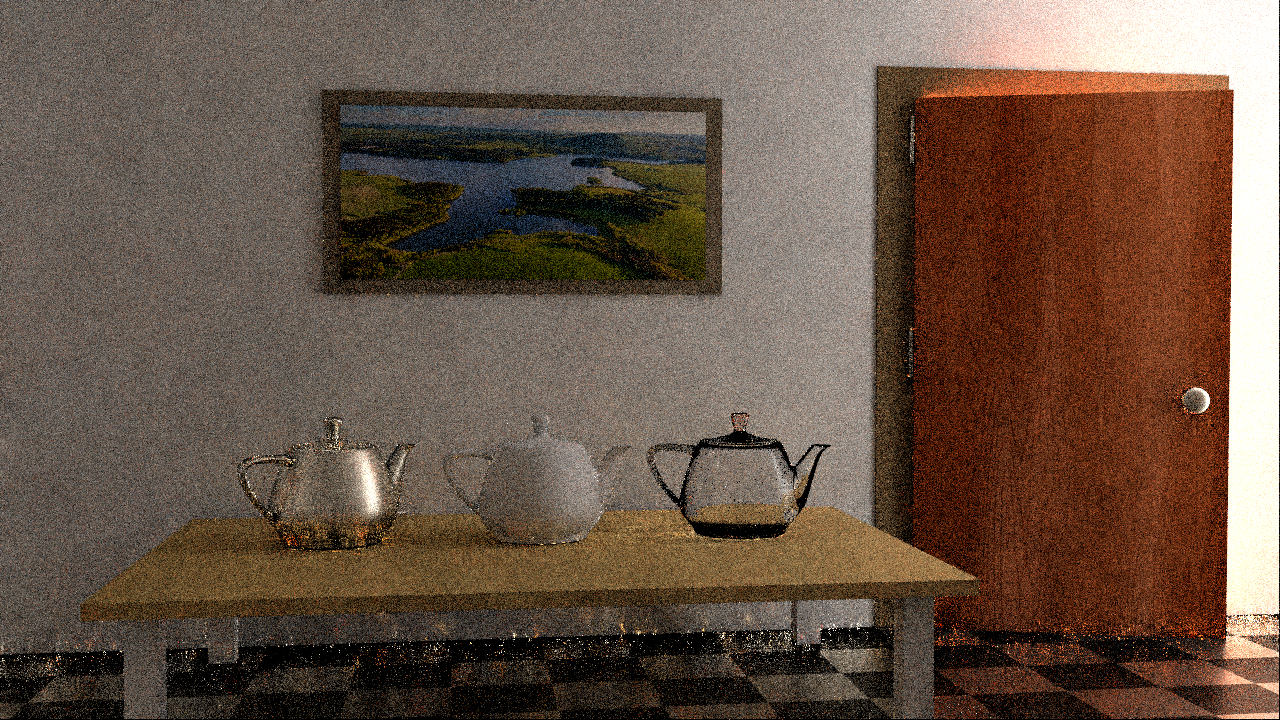} \\
\begin{tabular}{lll}
\includegraphics[width=0.111\textwidth]{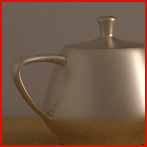} & \includegraphics[width=0.111\textwidth]{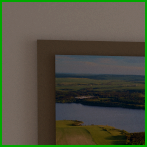} & \includegraphics[width=0.111\textwidth]{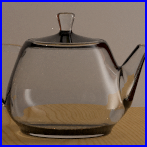}
\end{tabular} &
\begin{tabular}{lll}
\includegraphics[width=0.111\textwidth]{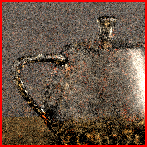} & \includegraphics[width=0.111\textwidth]{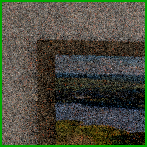} & \includegraphics[width=0.111\textwidth]{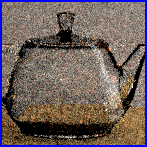}
\end{tabular} &
\begin{tabular}{lll}
\includegraphics[width=0.111\textwidth]{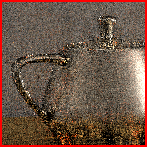} & \includegraphics[width=0.111\textwidth]{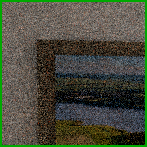} & \includegraphics[width=0.111\textwidth]{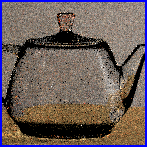}
\end{tabular} \\
\includegraphics[width=0.333\textwidth]{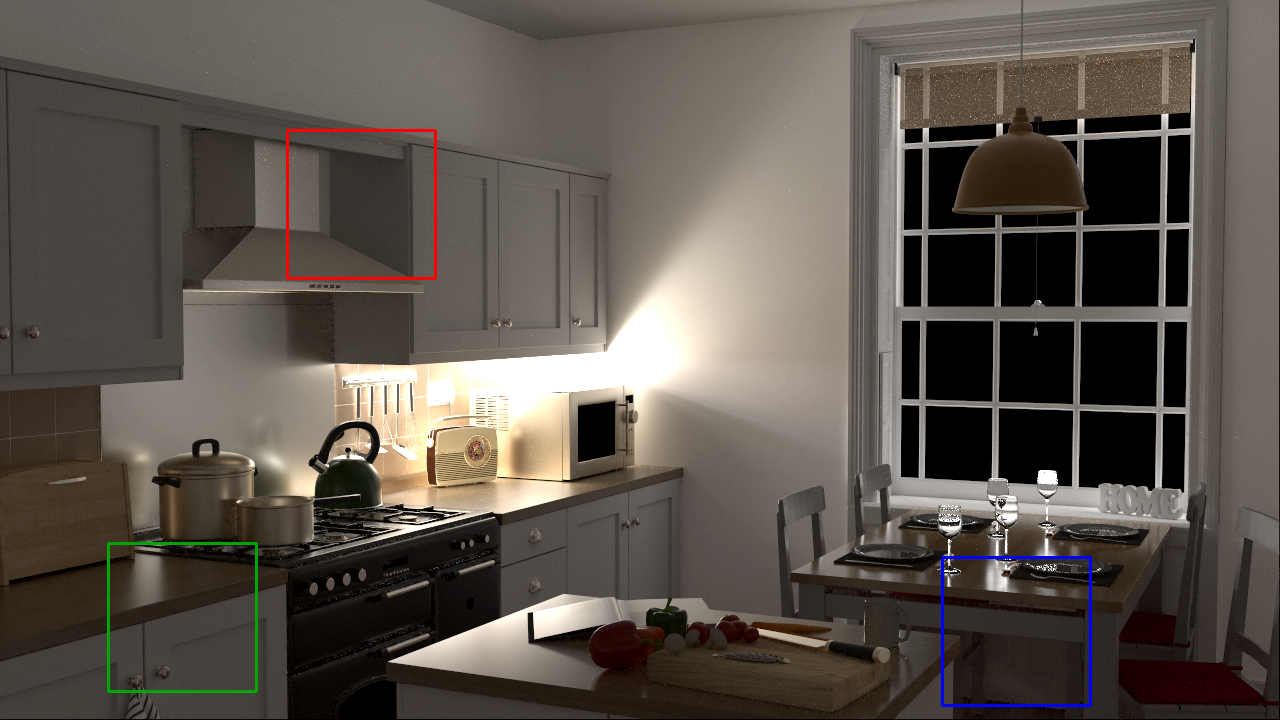} & \includegraphics[width=0.333\textwidth]{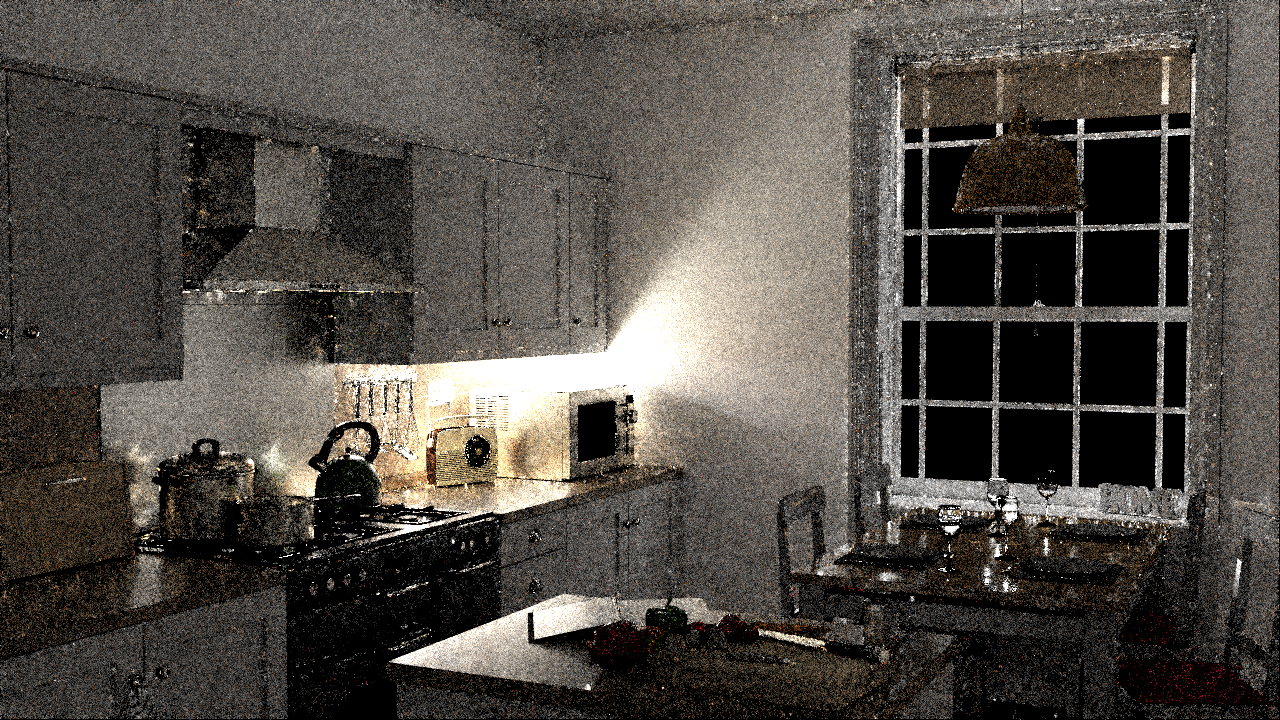} & \includegraphics[width=0.333\textwidth]{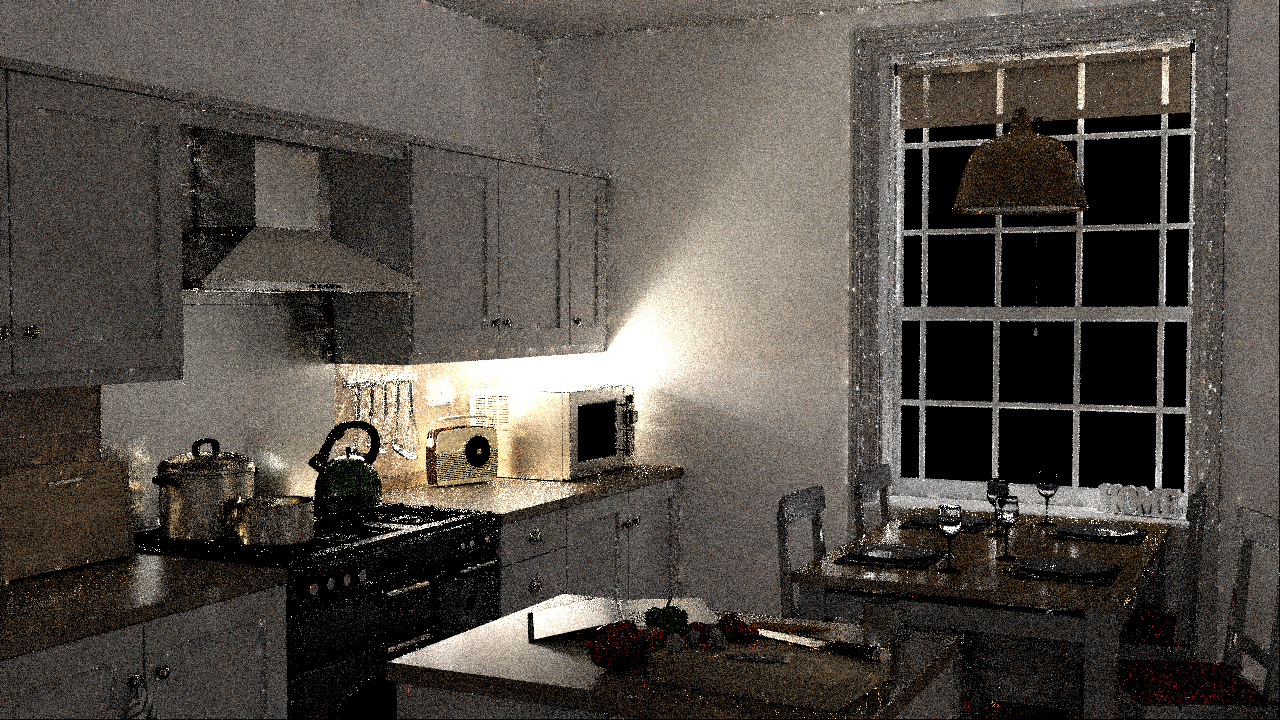} \\
\begin{tabular}{lll}
\includegraphics[width=0.111\textwidth]{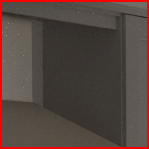} & \includegraphics[width=0.111\textwidth]{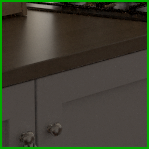} & \includegraphics[width=0.111\textwidth]{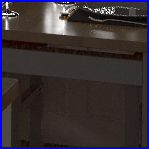}
\end{tabular} &
\begin{tabular}{lll}
\includegraphics[width=0.111\textwidth]{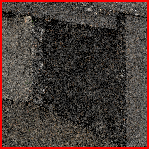} & \includegraphics[width=0.111\textwidth]{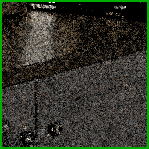} & \includegraphics[width=0.111\textwidth]{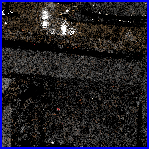}
\end{tabular} &
\begin{tabular}{lll}
\includegraphics[width=0.111\textwidth]{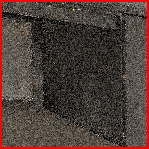} & \includegraphics[width=0.111\textwidth]{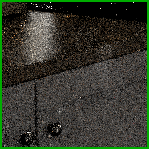} & \includegraphics[width=0.111\textwidth]{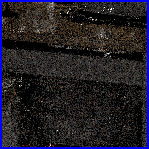}
\end{tabular} \\
Reference & MLT & Ours
\end{tabular}
\caption{Results for the \emph{door-ajar} and \emph{kitchen} scenes showing the reference (left) MLT (middle), and our method (right). Our approach reduces variance by distributing samples between partitions and the use of guided image perturbations.}
\label{fig:results}
\end{figure*}

Figure \ref{fig:results} shows the \emph{door-ajar} and \emph{kitchen} scenes which have an RMSE of 0.0811 (Ours) vs 0.0924 (MLT) and 0.0222 (Ours) vs 0.0225 (MLT) for the scenes respectively. Our approach leads to better convergence by distributing samples between partitions, as is highlighted by the unbalanced sampling of glass teapot in MLT in the \emph{door-ajar} scene at this low sample number compared to our method which leads to more equal coverage of the partitions.

\begin{figure}[htp]
\centering
\setlength{\tabcolsep}{0pt}
\renewcommand{\arraystretch}{0}
\begin{tabular}{lcccc}
& 8 & 32 & 64 & 128 \\
\rotatebox{90}{8} & \includegraphics[width=0.12\textwidth]{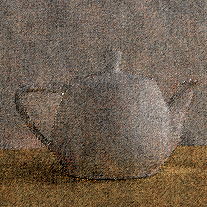} & \includegraphics[width=0.12\textwidth]{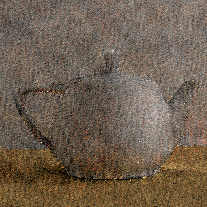} & \includegraphics[width=0.12\textwidth]{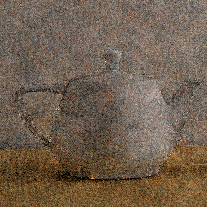} & \includegraphics[width=0.12\textwidth]{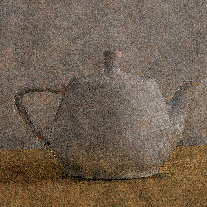} \\
\rotatebox{90}{24} & \includegraphics[width=0.12\textwidth]{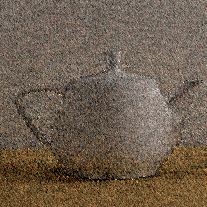} & \includegraphics[width=0.12\textwidth]{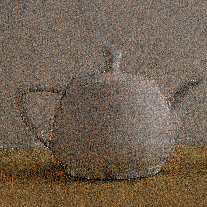} & \includegraphics[width=0.12\textwidth]{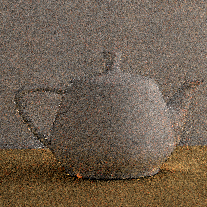} & \includegraphics[width=0.12\textwidth]{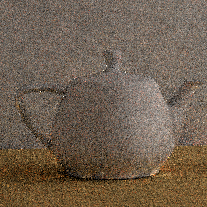} \\
\rotatebox{90}{44} & \includegraphics[width=0.12\textwidth]{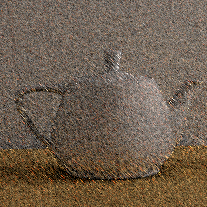} & \includegraphics[width=0.12\textwidth]{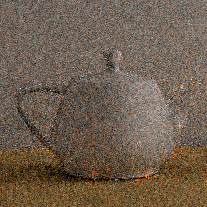} & \includegraphics[width=0.12\textwidth]{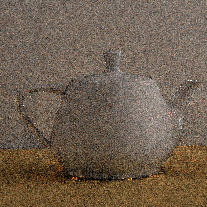} & \includegraphics[width=0.12\textwidth]{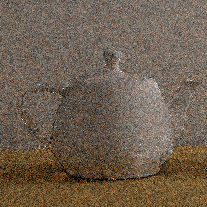} \\
\rotatebox{90}{128} & \includegraphics[width=0.12\textwidth]{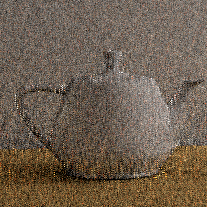} & \includegraphics[width=0.12\textwidth]{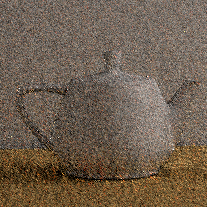} & \includegraphics[width=0.12\textwidth]{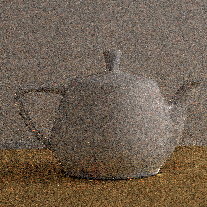} & \includegraphics[width=0.12\textwidth]{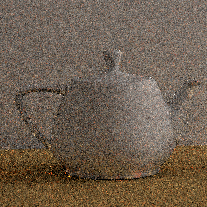}
\end{tabular}
\caption{Impact of the size and radius of the points in $Y'$. The rows show different radii, and the columns show the numbers of points.}
\label{fig:kernels}
\end{figure}

Finally, we show the impact of different sizes of $|Y'|$ and different with different radii in Figure \ref{fig:kernels}, which shows that too few points in $|Y'|$ leads to structured noise, and a too large radius leads to more high frequency noise.

Overall the results show that our method leads to an overall reduction in variance, however there can be increased noise compared to MLT in local regions of the scene. This is due to the sparse nature of $Y'$, and future work will investigate a more optimal set of values for $Y'$ to reduce this variance.

\section{Conclusion}

This paper proposed a principled approach to partition path space into a discrete set of subspaces. We propose an automated algorithm to choose these partitions, and an image space MCMC proposal distribution which can explore these spaces efficiently. This distribution is based on an analysis of terms in the acceptance probability and can be efficiently approximated via the use of denoising the initial samples used to create the partitions. Our results show an improvement in variance for the same number of samples across several scenes.

\bibliographystyle{ACM-Reference-Format}
\bibliography{References}

\end{document}